\documentclass[runningheads]{llncs}
\usepackage{graphicx}
\usepackage{comment}
\usepackage{amsmath,amssymb} 
\usepackage{color}

\usepackage{epstopdf}
\usepackage{tabularx,threeparttable}
\usepackage{array}
\usepackage[hidelinks]{hyperref}
\usepackage{graphicx}
\usepackage{subfigure}
\usepackage{amssymb, amsmath, bm}
\usepackage{mathtools}
\usepackage{mathrsfs}
\usepackage{booktabs}
\usepackage{multirow}
\usepackage{cite}
\usepackage{url}
\usepackage{color}
\usepackage{dsfont}

\newcommand\T{\rule{0pt}{2.4ex}}       
\newcommand\B{\rule[-1.2ex]{0pt}{0pt}} 
\newcommand{\tabincell}[2]{\begin{tabular}{@{}#1@{}}#2\end{tabular}}  
\def\ie{\emph{i.e.}}
\def\eg{\emph{e.g.}}
\def\etal{{\em et al.}}

\begin{document}
	\pagestyle{headings}
	\mainmatter
	\def\ECCVSubNumber{759}  
	
	\title{Learning from Extrinsic and Intrinsic Supervisions for Domain Generalization}

	\titlerunning{EISNet}
	%
	\author{Shujun Wang\inst{1}\orcidID{0000-0003-1495-3278} \and
	Lequan Yu\inst{2}\thanks{Corresponding Author}\orcidID{0000-0002-9315-6527} \and \\
	Caizi Li\inst{3} \and
	Chi-Wing Fu\inst{1,3}\orcidID{0000-0002-5238-593X}\and \\
	Pheng-Ann Heng\inst{1,3}\orcidID{0000-0003-3055-5034}}
	\authorrunning{S. Wang et al.}
	%
	\institute{The Chinese University of Hong Kong \\ \email{\{sjwang,cwfu,pheng\}@cse.cuhk.edu.hk} \and
		Stanford University \\ \email{lequany@stanford.edu} \and
		Guangdong Provincial Key Laboratory of Computer Vision and Virtual Reality Technology, Shenzhen Institutes of Advanced Technology, \\Chinese Academy of Sciences, Shenzhen, China \\
	\email{cz.li@siat.ac.cn}
}
	\maketitle

	\begin{abstract}
	The generalization capability of neural networks across domains is crucial for real-world applications.
	We argue that a generalized object recognition system should well understand the relationships among different images and also the images themselves at the same time.
	To this end, we present a new domain generalization framework (called EISNet) that learns how to generalize across domains \textit{simultaneously} from \textit{extrinsic relationship supervision} and \textit{intrinsic self-supervision} for images from multi-source domains.
	To be specific, we formulate our framework with feature embedding using a multi-task learning paradigm.
	Besides conducting the common supervised recognition task, we seamlessly integrate a momentum metric learning task and a self-supervised auxiliary task to collectively integrate the extrinsic and intrinsic supervisions.
	%
	Also, we develop an effective momentum metric learning scheme with the $K$-hard negative mining to boost the network generalization ability.
%
	%
	We demonstrate the effectiveness of our approach on two standard object recognition benchmarks VLCS and PACS, and show that our EISNet achieves state-of-the-art performance.

	
	
\end{abstract}

	\section{Introduction}

The rise of deep neural networks has achieved promising results in various computer vision tasks.
%
%
Most of these achievements are based on supervised learning, which assumes that the models are trained and tested on the samples drawn from the same distribution or domain.
However, in many real-world scenarios, the training and test samples are often acquired under different criteria.
Therefore, the trained network may perform poorly on ``unseen" test data with domain discrepancy from the training data.
To address this limitation, researchers have studied how to alleviate the performance degradation of a trained network among different domains.
For instance, by utilizing labeled (or unlabeled) target domain samples, various domain adaptation methods have been proposed to minimize the domain discrepancy by aligning the source and target domain distributions \cite{kumar2010co,ganin2016domain,hoffman2017cycada,saito2018maximum,tzeng2015simultaneous,tzeng2017adversarial}. 

Although these domain adaptation methods can achieve better performance on the target domain, there exists an indispensable demand to pre-collect and access target domain data during the network training. 
Moreover, it needs to re-train the network to adapt to every new target domain.
However, in real-world applications, it is often the case that adequate target domain data is not available during the training process~\cite{li2018domain,yue2019domain}.
For example, it is difficult for an automated driving system to know which domain (\eg, city, weather) the self-driving car will be used. 
Therefore, it has a broad interest in studying how to learn a generalizable network that can be directly applied to new ``unseen" target domains. 
Recently, the community develops \textit{domain generalization} methods to improve the model generalization ability on unseen target domains by utilizing the multiple source domains.

Most existing domain generalization methods attempt to extract the shared domain-invariant semantic features among multiple source domains~\cite{motiian2017unified,li2018domain,li2019episodic, li2018learning,dou2019domain}.
For example, Li~\etal~\cite{li2018domain} extend an adversarial auto-encoder by imposing the Maximum Mean Discrepancy (MMD) measure to align the distributions among different domains.
Since there is no specific prior information from target domains during the training, some works have investigated the effectiveness of increasing the diversity of the inputs by creating synthetic samples to improve the generalization ability of networks~\cite{yue2019domain,zakharov2019deceptionnet}. 
For instance, Yue~\etal~\cite{yue2019domain} propose a domain randomization method with Generative Adversarial Networks (GANs) to learn a model with high generalizability.
Meta-learning has also been introduced to address the domain generalization problem via an episodic training~\cite{li2019episodic, dou2019domain}.
Very recently, Carlucci~\etal~\cite{carlucci2019domain} introduce a self-supervision task by predicting relative positions of image patches to constrain the semantic feature learning for domain generalization.
This shows that the self-supervised task can discover invariance in images with different patch orders and thus improve the network generalization.
Such self-supervision task only considers the regularization within images but does not explore the valuable relationship among images across different domains to further enhance the discriminability and transferability of semantic features.

The generalization of deep neural networks relies crucially on the ability to learn and adapt knowledge across various domains.
%
%
%
We argue that a generalized object recognition system should well understand the relationships among different objects and the objects themselves at the same time.
Particularly, on the one hand, exploring the relationship among different objects (\ie, \textit{extrinsic supervision}) guides the network to extract domain-independent yet category-specific representation, facilitating decision-boundary learning.
On the other hand, exploring context or shape constraint within a single image (\ie, \textit{intrinsic supervision}) introduces necessary regularization for network training, broadening the network understanding of the object.
%

\begin{figure}[t]
	\centering
	\includegraphics[width=\textwidth]{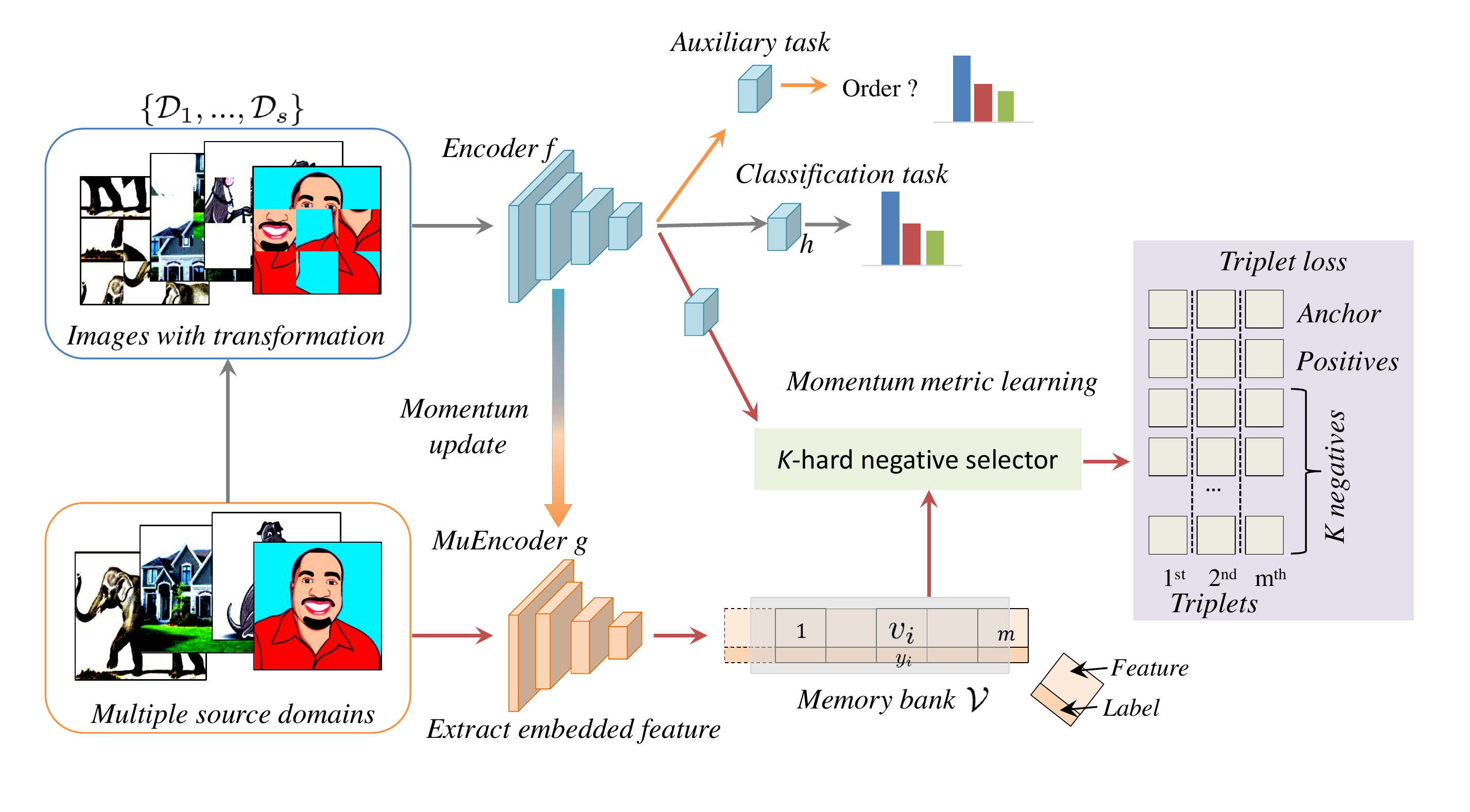}
	\caption{
		The framework of the proposed EISNet for domain generalization.
		We train a feature Encoder $f$ for discriminative and transferable feature extraction and a classifier for object recognition.
		Two complementary tasks, a momentum metric learning task and a self-supervised auxiliary task, are introduced to prompt general feature learning.
		We maintain a momentum updated Encoder (MuEncoder) to generate momentum updated embeddings stored in a large memory bank. Also, we design a $K$-hard negative selector to locate the informative hard triplets from the memory bank to calculate the triplet loss.
		The auxiliary self-supervised task predicts the order of patches within an image.}
	\label{fig:framework}
\end{figure}

To this end, we present a new framework called EISNet that learns how to generalize across domains by simultaneously incorporating \textit{extrinsic supervision} and \textit{intrinsic supervision} for images from multi-source domains.
%
We formulate our framework as a multi-task learning paradigm for general feature learning, as shown in Fig.~\ref{fig:framework}. 
Besides conducting the common supervised recognition task, we seamlessly integrate a momentum metric learning task and a self-supervised auxiliary task into our framework to utilize the \textit{extrinsic} and \textit{intrinsic} supervisions, respectively.
%
%
Specifically, we develop an effective momentum metric learning scheme with the $K$-hard negative selector to encourage the network to explore the image relationship and enhance the discriminability learning.
The $K$-hard negative selector is able to filter the informative hard triplets, while the momentum updated encoder guarantees the consistency of embedded features stored in the memory bank, which stabilizes the training process.
We then introduce a jigsaw puzzle solving task to learn the spatial relationship of images parts.
%
The three kinds of tasks share the same feature encoder and are optimized in an end-to-end manner.
We demonstrate the effectiveness of our approach on two object recognition benchmarks.
Our EISNet achieves the state-of-the-art performance.

	\section{Related Work}

\subsubsection{Domain adaptation and generalization}
The goal of unsupervised domain adaptation is to learn a general model with source domain images and unlabeled target domain images, so that the model could perform well on the target domain.
Under such a problem setting, images from the target domain can be utilized to guide the optimization procedure.
The general idea of domain adaption is to align the source domain and target domain distributions in the input level~\cite{hong2018conditional,chen2019synergistic}, semantic feature level~\cite{ren2018adversarial,dou2018unsupervised}, or output space level~\cite{chen2018road,tsai2018learning,vu2019advent,tsai2019domain,wang2019patch}.
Most methods adopt Generative Adversarial Networks and achieve better performance on the target domain data. 
However, training domain adaptation models need to access unlabeled target domain data, making it impractical for some real-world applications.

Domain generalization is an active research area in recent years. Its goal is to train a neural network on multiple source domains and produce a trained model that can be applied directly to unseen target domain.
Since there is no specific prior guidance from the target domain during the training procedure, some domain generalization methods proposed to generate synthetic images derived from the given multiple source domains to increase the diversity of the input images, so that the network could learn from a larger data space~\cite{zakharov2019deceptionnet,yue2019domain}.
Another promising direction is to extract domain-invariant features over multiple source domains~\cite{motiian2017unified,li2018domain,li2019episodic, li2018learning,ghifary2015domain}.
For example, Li~\etal~\cite{li2017deeper}~developed a low-rank parameterized CNN model for domain generalization and proposed the domain generalization benchmark dataset PACS.
Motiian~\etal~\cite{motiian2017unified} presented a unified framework by exploiting
the Siamese architecture to learn a discriminative space.
A novel framework based on adversarial autoencoders was presented by Li \etal ~\cite{li2018domain} to learn a generalized latent feature representation across domains.
Recently, meta-learning-based episodic training was designed to tackle domain generalization problems~\cite{li2019episodic,dou2019domain}.
Li~\etal~\cite{li2019episodic} developed an episodic training procedure to expose the network to domain shift that characterizes a novel domain at runtime to improve the robustness of the network.
Our work is most related to~\cite{carlucci2019domain}, which introduced self-supervision signals to regularize the semantic feature learning.
However, besides the self-supervision signals within a single image, we further exploit the extrinsic relationship among image samples across different domains to improve the feature compactness.

\subsubsection{Metric learning}
Our work is also related to metric learning, which aims to learn a metric to minimize the intra-class distances and maximize the inter-class variations~\cite{hadsell2006dimensionality, weinberger2009distance}.
With the development of deep learning, distance metric also benefits the feature embedding learning for better discrimination~\cite{wang2014learning,hoffer2015deep}.
Recently, the metric learning strategies have attracted a lot of attention on face verification and recognition \cite{schroff2015facenet}, fine-grained object recognition~\cite{wang2014learning}, image retrieval \cite{yuan2017hard}, and so on.
Different from previous applications, in this work, we adopt the conventional triplet loss with more informative negative selection and momentum feature extraction for domain generalization.

\subsubsection{Self-supervision}
Self-supervision is a recent paradigm for unsupervised learning. The idea is to design annotation-free (\ie, self-supervised) tasks for feature learning to facilitate the main task learning.
Annotation-free tasks can be predictions of the image colors~\cite{larsson2016learning}, relative locations of patches from the same image~\cite{noroozi2016unsupervised,carlucci2019domain}, image inpainting~\cite{pathak2016context}, and image rotation~\cite{gidaris2018unsupervised}.
Typically, self-supervised tasks are used as network pre-train to learn general image features. 
Recently, it is trained as an auxiliary task to promote the mainstream task by sharing semantic features~\cite{chen2019self}.
%
In this paper, we inherit the advantage of self-supervision to boost the network generalization ability.


	\section{Method}
We aim to learn a model that can perform well on ``unseen" target domain by utilizing multiple source domains.
Formally, we consider a set of $S$ source domains $\{\mathcal{D}_1, ..., \mathcal{D}_s\}$, with the $j$-th domain $\mathcal{D}_j$ having $N_j$ sample-label pairs $\{(x^j_i, y^j_i)\}^{N_j}_{i=1}$, where $x^j_i$ is the $i$-th sample in $\mathcal{D}_j$ and $y^j_i \in \{1,2,..., C\}$ is the corresponding label. 
In this work, we consider the object recognition task and aim to learn an Encoder $f_\theta : \ \mathclap{\mathcal{X}} \rightarrow \mathcal{Z}$ mapping an input sample $x_i$ into the feature embedding space $f_\theta(x_i) \in \mathcal{Z}$, where $\theta$ denotes the parameters of Encoder $f_\theta$.
We assume that Encoder $f_\theta$ could extract discriminative and transferable features, so that the task network (\eg, classifier) $h_\psi : \mathcal{Z} \rightarrow \mathbb{R}^C$ can be prompted on the unseen target domain. 

The overall framework of the proposed EISNet is illustrated in Fig.~\ref{fig:framework}.
We adopt the classical classification loss, \ie, Cross-Entropy, to minimize the objective $\mathcal{L}_c(h_\psi(f_\theta(x)),y)$ that measures the difference between the ground truth $y$ and the network prediction $\hat{y}=h_\psi(f_\theta(x))$.
%
%
To avoid performance degradation on unseen target domain,
we introduce two additional complementary supervisions to our framework.
One is an extrinsic supervision with momentum metric learning, and the other is an intrinsic supervision with a self-supervised auxiliary task.
The momentum metric learning is employed by a triplet loss with a $K$-hard negative selector on the momentum updated embeddings stored in a large memory bank.
We implement a self-supervised auxiliary task by predicting the order of patches within an image.
%
%
All these tasks adopt a shared encoder $f$ and are seamlessly integrated into an end-to-end learning framework.
%
Below, we introduce the extrinsic supervision and intrinsic self-supervision in detail.

\subsection{Extrinsic Supervision with Momentum Metric Learning}
\label{sec:relationship}

For the domain generalization problem, it is necessary to ensure the features of samples with the same label close to each other, while the features of different class samples being far apart.
Otherwise, the predictions on the unseen target domain may suffer from ambiguous decision boundaries and performance degradation~\cite{kamnitsas2018semi,dou2019domain}.
%
This is well aligned with the philosophy of metric learning. 
Therefore, we design a momentum metric-learning scheme to encourage the network to learn such domain-independent yet class-specific features by considering the mutual relation among samples across domains. 
Specifically, we propose a novel $K$-hard negative selector for triple loss to improve the training effectiveness by selecting informative triplets in the memory bank, and a momentum updated encoder to guarantee the representation consistency among the embeddings stored in the memory bank.

\begin{figure}[t]
	\centering
	\includegraphics[width=0.9\textwidth]{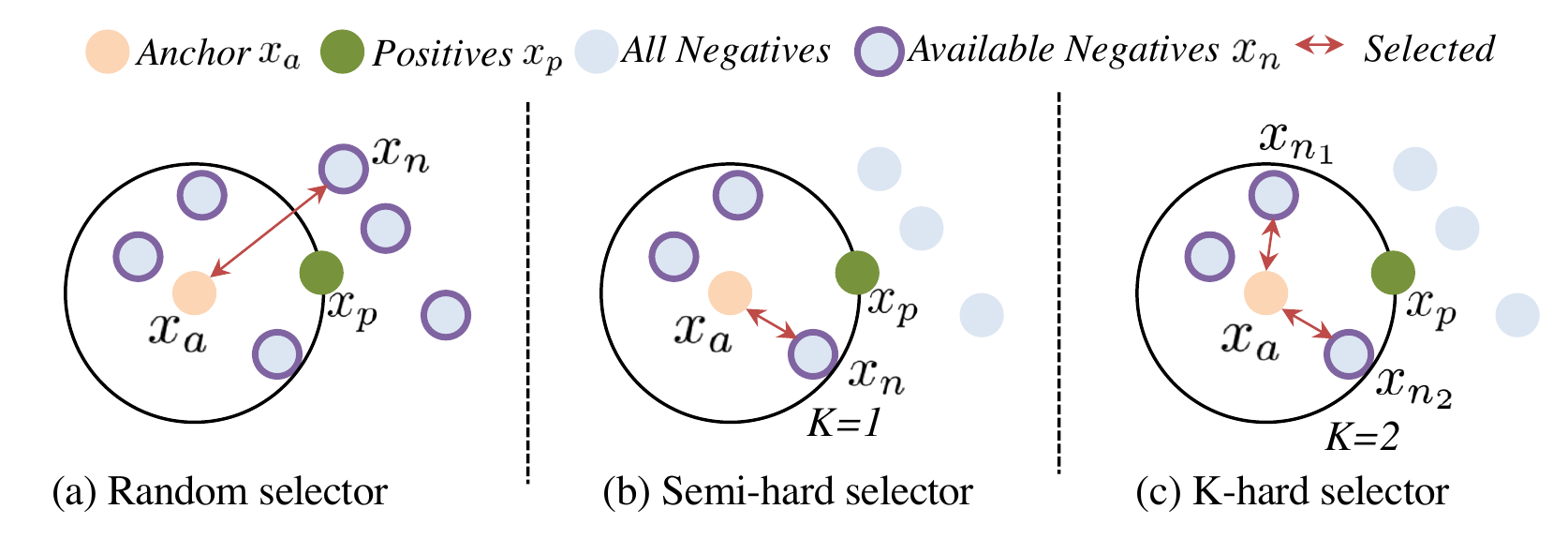}
	\caption{
		The schematic diagram of triplet negative sample selectors.
		We draw a circle with the anchor ($x_a$) as the center and the distance between the anchor ($x_a$) and positive ($x_p$) as the radius. We ignore the relaxation margin here and set $K$ as 2 for illustration. The selected negatives ($x_n$) are shown with arrows.
	}
	\label{fig:triplet}
\end{figure}

\subsubsection{$K$-hard negative selector for triplet loss}
The triplet loss is widely used to learn feature embedding based on the relative similarity of the sampled pairs.
%
%
The goal of the original triplet loss is to assign close distance to pairs of similar samples (\ie, positive pair) and long distance to pairs of dissimilar samples (\ie, negative pair).
For example, we can extract the feature representation $v_i$ of each image $x_i$ from multi-source domains with the feature Encoder $f_\theta$. 
Then by fixing an anchor sample $x_a$, we choose a corresponding positive sample $x_p$ with the same class label as $x_a$, and a random negative sample $x_n$ with different class label from $x_a$ to form a triplet $\mathcal{T}=\{(x_a, x_p,x_n)| y_a=y_p, \ y_a\neq y_n\}$.
Accordingly, the objective of the original triplet loss is formulated as 
\begin{align}
\mathcal{L}_\mathcal{T}=[d(x_a, x_p)^2 - d(x_a,x_n)^2 + \text{margin}]_+,
\end{align}
where $[\cdot ]_+ =max(0, \cdot)$, $d(x_i, x_j)$ represents the distance between the samples, and the margin is a standard relaxation coefficient. %
In general, we use the Euclidean distance to measure distances between the embedded features.
Then, the distance between samples $x_i$ and $x_j$ is defined as
\begin{align}
d(x_i, x_j) = \sqrt{\parallel v_i - v_j\parallel ^2 } =\sqrt{\parallel f_\theta(x_i) -  f_\theta(x_j) \parallel ^2 }.
\end{align}
%

The negative sample selection process in the original triplet loss is shown in Fig.~\ref{fig:triplet}(a).
Since the selected negative sample may already obey the triplet constraint, the training with the original triplet loss selector may not be efficient. 
To avoid useless training, inspired by~\cite{sohn2016improved,schroff2015facenet}, we propose a novel $K$-hard negative online selector, which extends the triplet with $K$ negatives that violate the triplet constraint within a certain of margin.
%
Specifically, given a sampled anchor, we randomly choose one positive sample with the same class label as the anchor, and select $K$ hard negative samples $x_{n_i}, i=\{1,2,...,K\}$ following
\begin{align}
\mathcal{T}'=\{(x_a, x_p,x_{n_i})| y_a=y_p, y_a\neq y_{n_i},  d(x_a,x_{n_i})^2 < d(x_a, x_p)^2+\text{margin}\}.
\end{align}
In an extreme case, the number of hard negative samples may be zero, then we random select negative samples without the distance constraint.
Therefore, the objective of the proposed triple loss with $K$-hard negative selector can be represented as
\begin{align}
\label{eq:triplet}
\mathcal{L}_{\mathcal{T}'}=\frac{1}{K} \sum_{i=1}^{K}[d(x_a, x_p)^2 - d(x_a,x_{n_i})^2 + \text{margin}]_+.
\end{align}
We illustrate the triplet selection process of semi-hard selector ($K=1$) and $K$-hard selector ($K=2$) in Fig.~\ref{fig:triplet} (b) (c) for a better understanding.
%
Compared with the original triplet loss, our proposed triple loss equipped with the $K$-hard negative selector considers more informative hard negatives for each anchor, thus facilitating the feature encoder to learn more discriminative features.

\subsubsection{Efficient learning with memory bank}
The way to select informative triplet pairs has a large influence on the feature embedding. 
Good features can be learned from a large sample pool that includes a rich set of negative samples~\cite{he2019momentum}.
However, selecting $K$-hard triplets from the whole sample pool is not efficient.
To increase the diversity of selected triplet pairs while reducing the computation burden, we maintain memory bank $\mathcal{V}$ to store the feature representation $v_i$ of historical samples~\cite{wu2018unsupervised} with a size of $m$.
Instead of calculating the embedded features of all the images at each iteration, we utilize the stored features to select the $K$-hard triplet samples.
Note that we also keep the class label $y_i$ along with representation $v_i$ in the memory bank to filter the negatives, as shown in Fig.~\ref{fig:framework}.
During the network training, we dynamically update the memory bank by discarding the oldest items and feeding the new batch of embedded features, where the memory bank acts as a queue.

\subsubsection{Momentum updated encoder}
With the memory bank, we can improve the efficiency of triplet sample selection. 
However, the representation consistency between the current samples and historical samples in the memory bank is reduced due to the rapidly-changed encoder~\cite{he2019momentum}.
Therefore, instead of utilizing the same feature encoder to extract the representation of current samples and historical samples, we adopt a new \textbf{M}omentum \textbf{u}pdated \textbf{E}ncoder (MuEncoder) to generate feature representation for the samples in the memory bank.
Formally, we denote the parameters of Encoder and MuEncoder as $\theta _f$ and $\theta _g$, respectively.
The Encoder parameter $\theta_f$ is optimized by a back-propagation of the loss function, while the MuEncoder parameter $\theta_g$ is updated as a moving average of Encoder parameters $\theta _f$ following
\begin{equation}
\theta _g = \delta * \theta _g + (1-\delta) * \theta _f, \ \  \text{where}\ \delta \in [0,1),
\end{equation}
where $\delta$ is a momentum coefficient to control the update degree of MuEncoder.
Since the MuEncoder evolves more smoothly than Encoder, the update of different features in the memory bank is not rapid, thereby easing the triplet loss update.
This is confirmed by the experimental results.
%
In our preliminary experiments, we found that a large momentum coefficient $\delta$ by slowly updating $\theta _g$ could generate better results than rapid updating, which indicates that a slow update of MuEncoder is able to guarantee the representation consistency.

\subsection{Intrinsic Supervision with Self-supervised Auxiliary Task}
\label{sec:selfsupervised}
To broaden the network understanding of the data, 
we propose to utilize the intrinsic supervision within a single image to impose a regularization into the feature embedding by adding auxiliary self-supervised tasks on all the source domain images.
%
%
A similar idea has been adopted in domain adaptation and Generative Adversarial Networks training~\cite{sun2019unsupervised,chen2019self}.
The auxiliary self-supervised task is able to exploit the intrinsic semantic information within a single image to provide informative feature representations for the main task.

There are plenty of works focusing on designing auxiliary self-supervised tasks, such as rotation degree prediction and relative location prediction of two patches in one image~\cite{doersch2015unsupervised,gidaris2018unsupervised,kolesnikov2019revisiting}.
Here, we employ the recently-proposed \textit{solving jigsaw puzzles}~\cite{noroozi2016unsupervised, carlucci2019domain} as our auxiliary task.
However, most of the self-supervised tasks focusing on high-level semantic feature learning can be incorporated into our framework. 
Specifically,  we first divide an image into nine ($3 \times 3$) patches, and shuffle these patches within the $30$ different combinations following \cite{carlucci2019domain}.
As pointed by~\cite{carlucci2019domain}, the model achieves the highest performance when the class number is set as $30$ and the order prediction performance decreases when the task becomes more difficult with more orders.
A new auxiliary task branch $h_a$ follows the extracted feature representation $f_\theta$ to predict the ordering of the patches.
A Cross-Entropy loss is applied to tackle this order classification task:
\begin{equation}
\mathcal L_{a} = - \frac{1}{N*31} \sum _{i=1}^{N} \sum _{c_{a}=0}^{30}{y^a_{i, c_a} * \text{log} (p^a_{i, c_a})},
\end{equation}
where $y^a$ and $p^a$ are the ground-truth order and predicted order from the auxiliary task branch, respectively. 
We use $c_a=0$ to represent the original images without patch shuffle, leading to a total of 31 classes.
%

%

Overall, we formulate the whole framework as a multi-task learning paradigm. The total objective function to train the network is represented as
\begin{equation}
\label{eq:total}
\mathcal{L} = \alpha * \mathcal{L}_c + \beta * \mathcal{L}_{\mathcal{T}'} +\gamma * \mathcal{L}_{a},
\end{equation}
where $\alpha,\ \beta,\ \text{and}\ \gamma$ are hyper-parameters to balance the weights of the basic classification supervision, extrinsic relationship supervision, and intrinsic self-supervision, respectively.

	\section{Experiments}

\subsection{Datasets}
We evaluate our method on two public domain generalization benchmark datasets: \textbf{VLCS} and \textbf{PACS}. \textbf{VLCS}~\cite{fang2013unbiased} is a classic domain generalization benchmark for image classification, which includes five object categories from four domains (PASCAL VOC 2007, LabelMe, Caltech, and Sun datasets).
\textbf{PACS}~\cite{li2017deeper} is a recent domain generalization benchmark for object recognition with larger domain discrepancy. It consists of seven object categories from four domains (Photo, Art Paintings, Cartoon, and Sketches datasets) and the domain discrepancy among different datasets is more severe than VLCS, making it more challenging. 

\subsection{Network Architecture and Implementation Details}
Our framework is flexible and one can use different network backbones as the feature Encoder. 
We utilized a fully-connected layer with $31$-dimensional output as the self-supervised auxiliary classification layer following the setting in \cite{carlucci2019domain} for a fair comparison. 
To enable the momentum metric learning, we further employed a fully-connected layer with $128$ output channels following the Encoder part and added an L2 normalization layer to normalize the feature representation $v$ of each sample.
The MuEncoder has the same network architecture as the Encoder, and the weight of MuEncoder was initialized with the same weight as Encoder.
We followed the previous works in the literature~\cite{balaji2018metareg,carlucci2019domain,li2018learning,dou2019domain} and employed the leave-one-domain-out cross-validation strategy to produce the experiment results, i.e., we take turns to choose each domain for testing, and train a network model with the remaining three domains.

We implemented our framework with the PyTorch library on one NVIDIA TITAN Xp GPU.
Our framework was optimized with the SGD optimizer. We totally trained 100 epochs, and the batch size was 128. 
The learning rate was set as 0.001 and decreased to 0.0001 after 80 epochs.
We empirically set the margin of the triplet loss as $2$.
We also adopted the same on-the-fly data augmentation as JiGen~\cite{carlucci2019domain}, which includes random cropping, horizontal flipping, and jitter.

\subsection{Results on VLCS Dataset}
We followed the same experiment setting in previous work~\cite{carlucci2019domain} to train and evaluate our method.
The extrinsic metric learning and intrinsic self-supervised learning was developed upon the ``FC7" features of AlexNet~\cite{NIPS2012_4824} pretrained on ImageNet~\cite{deng2009imagenet}.
We set the size of the memory bank as $1024$ and the number of negatives $K$ in the triplet loss Eq.~\eqref{eq:triplet} as $256$.
The hyper-parameters $\alpha$, $\beta$, and $\gamma$ in total objective function Eq.~\eqref{eq:total} were set as $1$, $0.1$, and $0.05$, respectively.
For our results, we report the average performance and standard deviation over three independent runs.

We compare our method with other nine previous state-of-the-art methods.
\textbf{D-MTAE} \cite{ghifary2015domain} utilized the multi-task auto-encoders to learn robust features across domains.
\textbf{CIDDG} \cite{li2018deep} was a conditional invariant adversarial network that learns the domain-invariant representations under distribution constraints.
\textbf{CCSA} \cite{motiian2017unified} exploited a Siamese network to learn a discriminative embedding subspace with distribution distances and similarities.
\textbf{DBADG} \cite{li2017deeper} developed a low-rank parametrized CNN model for domain generalization.
\textbf{MMD-AAE} \cite{li2018domain} aligned the distribution through an adversarial auto-encoder by Maximum Mean Discrepancy.
\textbf{MLDG} \cite{li2018learning} was a meta-learning method by simulating train/test domain shift during training.
\textbf{Epi-FCR} \cite{li2019episodic} was an episodic training method.
\textbf{JiGen} \cite{carlucci2019domain} solved a jigsaw puzzle auxiliary task based on self-supervision. 
\textbf{MASF} \cite{dou2019domain} employed a meta-learning based strategy with two complementary losses for encoder regularization.
Moreover, we include the \textbf{Within domain} performance of all the datasets as a comparison to reveal the performance drop due to domain discrepancy.
We trained \textbf{Within domain} using a supervised way with training and test images from the same domain.

The comparison results with the above methods are shown in Table~\ref{tab:results-VLCS}. 
%
%
It is observed that our EISNet achieves the best performance on both Caltech and Sun datasets and comparable results on PASCAL VOC and LabelMe datasets.
Overall, EISNet achieves an average accuracy of $74.67\%$ over four domains, outperforming the previous state-of-the-art method \textbf{MASF}~\cite{dou2019domain}.
%
%
Our method also outperforms \textbf{JiGen}~\cite{carlucci2019domain} on three domains and achieves comparable results on the remaining PASCAL VOC domain, demonstrating that utilizing extrinsic relationship supervision can further improve the network generalization ability. 
%

\begin{table*} [!t]
	\centering
	\caption{Domain generalization results on \textbf{VLCS} dataset with object recognition accuracy (\%) using \textbf{AlexNet} backbone. 
		The top results are highlighted in \textbf{bold}.}
	\label{tab:results-VLCS}
	\resizebox{1.0\textwidth}{!}{
		\setlength\tabcolsep{1.5pt}
		\begin{tabular}{l|cccccccccc|c}
			\toprule[1pt]
			\textbf{Target} & \tabincell{c}{\textbf{Within }\\\textbf{domain}} & \tabincell{c}{\textbf{D-MTAE} \\ \cite{ghifary2015domain}} & \tabincell{c}{\textbf{CIDDG}\\ \cite{li2018deep}} & \tabincell{c}{\textbf{CCSA}\\ \cite{motiian2017unified}} & \tabincell{c}{\textbf{DBADG}\\ \cite{li2017deeper}}& 
			\tabincell{c}{\textbf{MMD-}\\ \textbf{AAE} \cite{li2018domain} }&  \tabincell{c}{\textbf{MLDG} \\ \cite{li2018learning}} & \tabincell{c}{\textbf{Epi-}\\ \textbf{FCR}  \cite{li2019episodic}} & \tabincell{c}{\textbf{JiGen}\\ \cite{carlucci2019domain}} &  \tabincell{c}{\textbf{MASF}\\ \cite{dou2019domain}} &  \tabincell{c}{\textbf{EISNet} \\ {(Ours)}} \\ \hline
			\T
			PASCAL & 82.07 & 63.90 & 64.38 & 67.10 & 69.99 & 67.70 & 67.7 & 67.1 &\textbf{70.62} & 69.14 & 69.83$\pm$0.48  \\ 
			
			\textbf{L}abelMe & 74.32 & 60.13 & 63.06 & 62.10 & 63.49 & 62.60 & 61.3 & 64.3 &60.90 & \textbf{64.90 }& 63.49$\pm$0.82   \\ 
			\textbf{C}altech & 100.0 & 89.05& 88.83 & 92.30 & 93.63 & 94.40 & 94.4 & 94.1 &96.93 & 94.78 &\textbf{97.33}$\pm$0.36  \\     
			\textbf{S}un & 77.33 & 61.33 & 62.10 & 59.10 & 61.32 & 64.40 & 65.9 & 65.9 &64.30 & 67.64 &\textbf{68.02}$\pm$0.81  \\  \hline \T
			\textbf{Average} & 83.43 & 68.60& 69.59 & 70.15 & 72.11 & 72.28 & 72.3 & 72.9 &73.19 & 74.11 &\textbf{74.67}  \\ 
			\toprule[1pt]
		\end{tabular}
	}
\end{table*}

\begin{table*} [!t]
	\centering
	\caption{Domain generalization results on \textbf{PACS} dataset with object recognition accuracy (\%) using \textbf{AlexNet} backbone. 
		The top results are highlighted in \textbf{bold}.
	}
	\label{tab:results-PACS-alex}
	\resizebox{1.0\textwidth}{!}{
		\setlength\tabcolsep{1.5pt}
		\begin{tabular}{l|ccccccccc|c}
			\toprule[1pt]
			\textbf{Target} & \tabincell{c}{\textbf{Within }\\\textbf{domain}} & \tabincell{c}{\textbf{D-MTAE} \\ \cite{ghifary2015domain}} & \tabincell{c}{\textbf{CIDDG}\\ \cite{li2018deep}}  & \tabincell{c}{\textbf{DBADG}\\ \cite{li2017deeper}}&  \tabincell{c}{\textbf{MLDG} \\ \cite{li2018learning}} & \tabincell{c}{\textbf{Epi-}\\ \textbf{FCR} 
				\cite{li2019episodic}} &
			\tabincell{c}{\textbf{MetaReg} \\\cite{balaji2018metareg}} & \tabincell{c}{\textbf{JiGen}\\ \cite{carlucci2019domain}} &  \tabincell{c}{\textbf{MASF}\\ \cite{dou2019domain}} & \tabincell{c}{\textbf{EISNet} \\ {(Ours)}}  \\ \hline
			\T
			\textbf{P}hoto & 97.80 & 91.12 & 78.65     & 89.50 & 88.00 & 86.1 & 91.07 & 89.00 &90.68  & \textbf{91.20}$\pm$0.00  \\ 
			
			\textbf{A}rt painting & 90.36 & 60.27& 62.70 & 62.86 & 66.23 & 64.7 & 69.82 & 67.63 &70.35  & \textbf{70.38}$\pm$0.37   \\ 
			
			\textbf{C}artoon & 93.31 & 58.65& 69.73 & 66.97 & 66.88 & 72.3 & 70.35 & 71.71 &\textbf{72.46} & 71.59$\pm$1.32  \\     
			
			\textbf{S}ketch & 93.88 & 47.68& 64.45 & 57.51 & 58.96 & 65.0 & 59.26 & 65.18&67.33 & \textbf{70.25}$\pm$1.36   \\  \hline \T
			
			\textbf{Average} & 93.84 & 64.48& 68.88 & 69.21 & 70.01 & 72.0 & 72.62 &  73.38 & 75.21 &\textbf{75.86}    \\ 
			\toprule[1pt]
		\end{tabular}
	}
\end{table*}

\begin{table*} [!t]
	\centering
	\caption{Domain generalization results on \textbf{PACS} dataset with object recognition accuracy (\%) using \textbf{ResNet} backbones. 
		The top results are highlighted in \textbf{bold}.}
	\label{tab:results-PACS-resnet}
	\resizebox{1.0\textwidth}{!}{
		\setlength\tabcolsep{1.5pt}
		\begin{tabular}{l|ccc|ccc}
			\toprule[1pt]
			\multirow{2}{*}{\textbf{Target}} & \multicolumn{3}{c|}{\textbf{ResNet-18}} &
			\multicolumn{3}{c}{\textbf{ResNet-50}}\\
			\cline{2-7}  & \textbf{DeepAll} & \textbf{MASF} \cite{dou2019domain} & \tabincell{c}{\textbf{EISNet} {(Ours)}}  & \textbf{DeepAll} & \textbf{MASF} \cite{dou2019domain} & \tabincell{c}{\textbf{EISNet} {(Ours)}}  \T \B \\
			\hline
			\T
			\textbf{P}hoto & 94.25 & 94.99     &\textbf{95.93}$\pm$0.06  & 94.83 & 95.01  & \textbf{97.11}$\pm$0.40  \\ 
			
			\textbf{A}rt painting & 77.38& 80.29  &\textbf{81.89}$\pm$0.88 & 81.47 &82.89  & \textbf{86.64}$\pm$1.41   \\ 
			
			\textbf{C}artoon & 75.65& \textbf{77.17} & {76.44}$\pm$0.31 & 78.61&80.49 & \textbf{81.53}$\pm$0.64  \\     
			
			\textbf{S}ketch & 69.64&  71.69 & \textbf{74.33}$\pm$1.37 & 69.69&72.29 & \textbf{78.07}$\pm$1.43  \\  \hline \T
			
			\textbf{Average} & 79.23& 81.04& \textbf{82.15} &  81.15& 82.67&\textbf{85.84}   \\ 
			\toprule[1pt]
		\end{tabular}
	}
\end{table*}

\subsection{Results on PACS Dataset}

%
To show the effectiveness of our framework under different network backbones on PACS dataset, we evaluate our method with three different backbones: AlexNet, ResNet-18, and ResNet-50~\cite{He_2016_CVPR}.
%
%
The size of memory bank was set as $1024$ and $K$ in the triplet loss Eq.~\eqref{eq:triplet} was set as $256$.
The hyper-parameters in total objective function Eq.~\eqref{eq:total} were set as $1$, $0.5$, and $0.7$ for $\alpha$, $\beta$, and $\gamma$, respectively.
For our results, we also report the average performance and standard deviation over three independent runs.

Table~\ref{tab:results-PACS-alex} summarizes the experimental results developed with AlexNet backbone.
We compare our methods with eight other methods that achieved previous best results on this benchmark dataset.
\textbf{MetaReg} \cite{balaji2018metareg} utilized a novel classifier regularization in the meta-learning framework.
As we can observe from Table~\ref{tab:results-PACS-alex}, by simultaneously utilizing momentum metric learning and intrinsic self-supervision for images across different source domains, our method achieves the best performance on three datasets. 
Across all domains, our method achieves an average accuracy of $75.86\%$, setting a new state-of-the-art performance.

We also compare our method with baseline method (DeepAll) and the state-of-the-art method \textbf{MASF}~\cite{dou2019domain} using ResNet-18 and ResNet-50 backbones in  Table~\ref{tab:results-PACS-resnet}. 
In the ResNet-50 experiment, we reduce the batch size to 64 to fit the limited GPU memory.
The DeepAll method is trained with all the source domains without any specific network design.
As shown in Table~\ref{tab:results-PACS-resnet}, our method consistently outperforms \textbf{MASF} about 1.11\% and 3.17\% on average accuracy with ResNet-18 and ResNet-50 backbone, respectively.
This indicates that our designed framework is very general and can be migrated to different network backbones.
Note that the improvement over \textbf{MASF} is more obvious with a deeper network backbone, showing that our proposed algorithm is more beneficial for domain generalization with deeper feature extractors.

\begin{figure}[!t]
	\centering
	\includegraphics[width=0.9\textwidth]{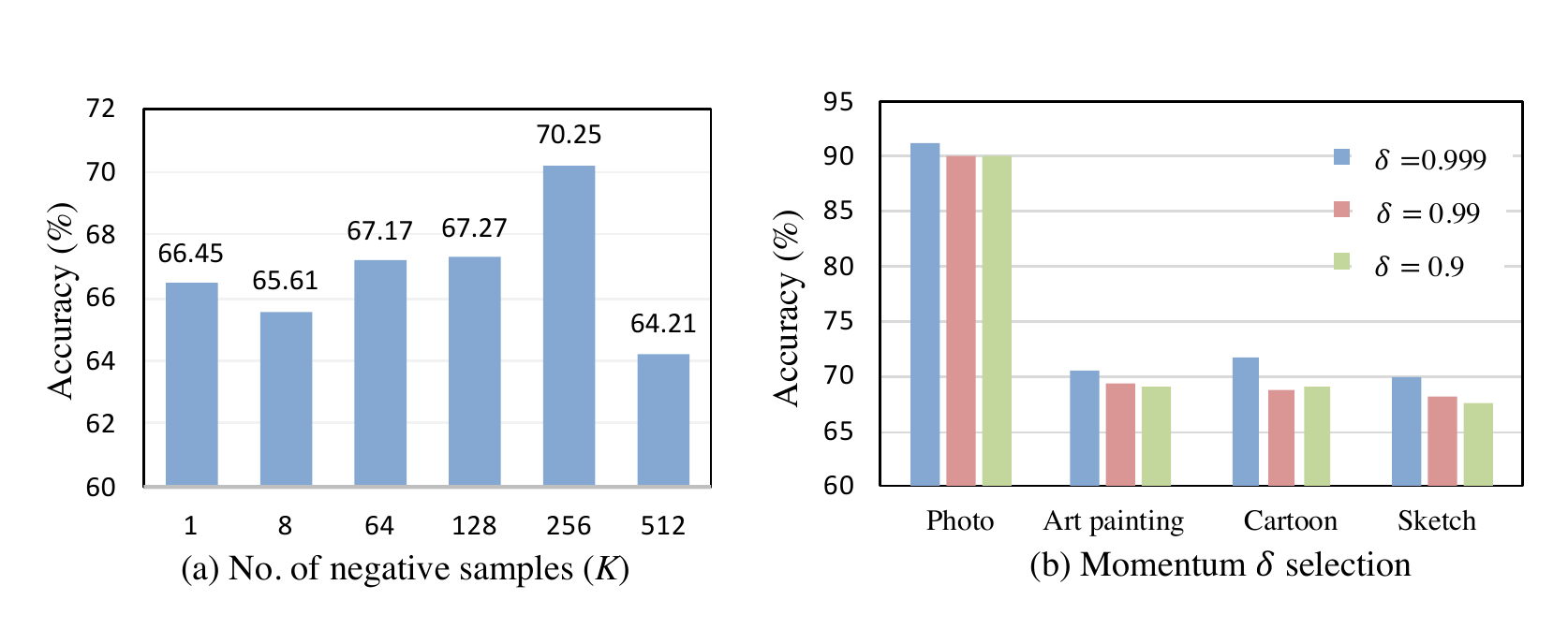}
	\caption{
		The performance of our method under different number of negative samples $K$ and momentum update coefficient $\delta$. 
	}
	\label{fig:ablation1}
\end{figure}
\begin{table*} [!t]
	\centering
	\caption{Ablation study on key components of our method with the \textbf{PACS} dataset (\%). 
		The top results are highlighted in \textbf{bold}.}
	\label{tab:results-PACS-ablation}
	{
		\setlength\tabcolsep{1.5pt}
		\begin{tabular}{cc|cccc|c}
			\toprule[1pt]
			\textbf{Extrinsic} & \textbf{Intrinsic}  & \textbf{P}hoto  & \textbf{A}rt painting &  \textbf{C}artoon &\textbf{S}ketch & \textbf{Average} \\
			\hline
			\T
			-&-&94.85&81.47&78.61&69.69& 81.15\\
			\checkmark &-&97.06&81.97&80.70&76.81&84.14 \\
			- &\checkmark&97.02&85.17&76.35&76.97&83.88 \\
			\checkmark     &\checkmark&\textbf{97.11}&\textbf{86.64}&\textbf{81.53}&\textbf{78.07}&\textbf{85.84} \\
			
			\toprule[1pt]
		\end{tabular}
	}
\end{table*}

\subsection{Analysis of Our Method}
We conduct extensive analysis of our method.  
Firstly, we investigate the effectiveness of extrinsic and intrinsic supervision using ResNet-50 backbone on \textbf{PACS} dataset, and the experimental results are illustrated in Table~\ref{tab:results-PACS-ablation}.
The \textbf{Extrinsic} supervision indicates that the momentum metric learning is used, while \textbf{Intrinsic} supervision denotes that the auxiliary self-supervision loss is optimized. 
The method without these two supervisions is the baseline model, which is the same with DeepAll results in Table~\ref{tab:results-PACS-resnet}.
From the results in Table~\ref{tab:results-PACS-ablation}, we observe that each supervision plays an important role in our framework.
Specifically, equipping the extrinsic supervision into the baseline model yields about 2.99\% average accuracy improvement.
Meanwhile, we also achieve 2.73\% average accuracy improvement over the baseline model by incorporating intrinsic self-supervision of the images.
By combing extrinsic and intrinsic supervision, performance is further improved across all settings, indicating these two supervisions are complementary.

\begin{table*} [!t]
	\centering
	\caption{Comparison of our proposed $K$-hard negative selector with original random selector and semi-hard negative selector.}
	\label{tab:results-ablation-selector}
	{
		\setlength\tabcolsep{1.5pt}
		\begin{tabular}{c|ccc}
			\toprule[1pt]
			\T
			Selector & Random & Semi-hard & $K$-hard \\ \hline \T
			Accuracy (\%) &65.38 &68.08&70.25   \\
			\toprule[1pt]
		\end{tabular}
	}
\end{table*}
\begin{table*} [!t]
	\centering
	\caption{Comparison among different memory bank size.}
	\label{tab:results-ablation-bank}
	{
		\setlength\tabcolsep{1.5pt}
		\begin{tabular}{c|cccc}
			\toprule[1pt]
			\T
			Memory bank size $m$ & 1024 & 512 & 256 & 128 \\ \hline
			No. of negatives $K$ & 256 & 128 & 64 & 32\\ \hline
			\T 
			Accuracy (\%) &70.25 &68.80&68.24 & 67.93   \\
			\toprule[1pt]
		\end{tabular}
	}
\end{table*}

We then analyze five key components in our framework, that is a) the number of different negative samples $K$ in momentum metric learning, b) the effectiveness of momentum update coefficient $\delta$, c) the effectiveness of hard negative selector, d) the size of memory bank $m$, and e) time cost.
All below comparison experiments are implemented with AlexNet backbone on the PACS benchmark.

\renewcommand{\theenumi}{\alph{enumi}}
\begin{enumerate}
	\item The number of negative samples $K$ is a key parameter of our designed $K$-hard negative selector in momentum metric learning. 
	We investigate the network performance under different options.
	We select six $K$ values at different magnitudes, which are 1, 8, 64, 128, 256, and 512.
	The Sketch dataset results are shown in Fig.~\ref{fig:ablation1} (a), 
	We can observe that a large number of negative samples would lead to better results in general and the network generates the best result with $K=256$. 
	However, the performance drops drastically if we set $K=512$, demonstrating that too large $K$ will produce a burden on the metric distance calculation and make the network difficult to learn.
	
	\item The momentum update coefficient $\delta$ is important to control the feature consistency among different batches of embedded features in the memory bank. 
	We show the accuracy with different momentum coefficient $\delta$ in Fig.~\ref{fig:ablation1} (b).
	It is observed that the network performs well when $\delta$ is relatively large, \ie, 0.999.
	A small coefficient would degrade the network performance, suggesting that a slow updating MuEncoder is beneficial to the feature consistency.
	
	\item To validate the effectiveness of $K$-hard negative selector in our proposed metric learning, we compare our proposed $K$-hard negative selector with original random triplet selector and semi-hard negative selector.
	The Sketch dataset results are shown in Table~\ref{tab:results-ablation-selector}.
	Equipped with semi-hard negative selector, the accuracy improves $2.70\%$.
	By selecting more negative pairs from the memory bank, we obtain the accuracy of $70.25\%$, demonstrating the effectiveness of the proposed $K$-hard negative selector.
	
	\item The size of memory bank $m$ can be adjusted according to different tasks. Here, we show the results of four different settings with the number of negatives changing as well in Table~\ref{tab:results-ablation-bank}. 
	In general, our method is able to generate better results with a large memory bank size and negative samples.
	However, a too large memory bank will increase the burden to calculate the pair-wise distance in triplet loss.
	Therefore, we need to balance the accuracy and computation burden.
	\item Apart from the performance improvement over other methods, our method has much lower computation cost. Under the same server setting (one TITAN XP GPU) and AlexNet backbone, our method only takes 1.5 hours to train the network on PACS dataset, while the total training time of the state-of-the-art MASF is about 17 hours. Therefore, our method could save more than $91\%$ time cost on training phase.
\end{enumerate}


\begin{figure}[!t]
	\centering
	\includegraphics[width=0.9\textwidth]{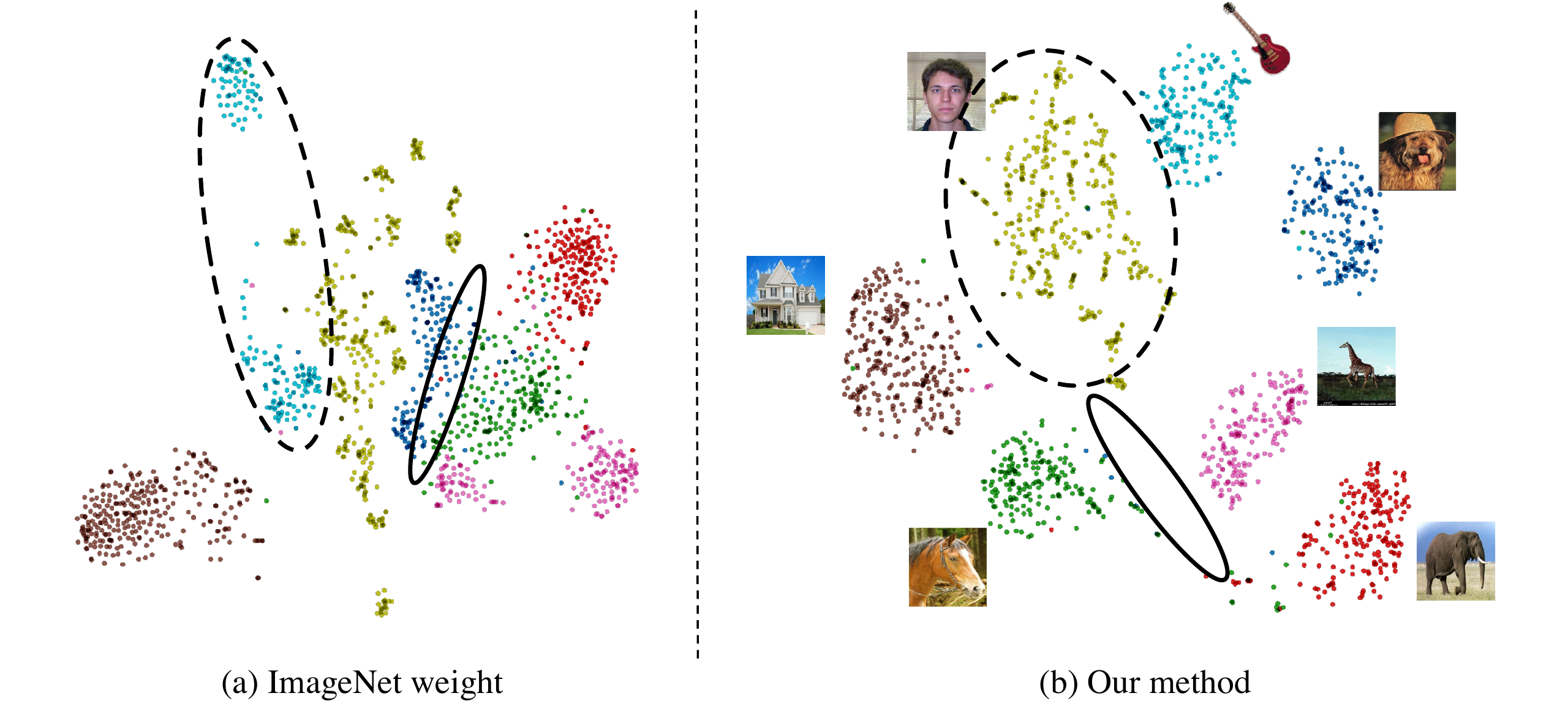}
	\caption{
		t-SNE visualization on one target domain to show the discrimination of the network. (a) is the feature embedding extracted from the IMAGENET pre-trained network. (b) shows the feature embedding distributions extracted from our EISNet.
	}
	\label{fig:tsne}
\end{figure}

We also employ t-SNE~\cite{maaten2008visualizing} to analyze the feature level discrimination of our method and the visualization results are shown in Fig.~\ref{fig:tsne}.
Compared with the feature extracted from the ImageNet pre-trained network, the distance between different class clusters in our method becomes evident, indicating that equipped with our proposed extrinsic and intrinsic supervision, the model is able to learn more discriminative features among different object categories regardless domains.

	\section{Conclusions}
We have presented a multi-task learning paradigm to learn how to generalize across domains for domain generalization.
The main idea is to learn a feature embedding simultaneously from the extrinsic relationship of different images and the intrinsic self-supervised constraint within the single image. 
We design an effective and efficient momentum metric learning module to facilitate compact feature learning.   
Extensive experimental results on two public benchmark datasets demonstrate that our proposed method is able to learn discriminative yet transferable feature, which lead to state-of-the-art performance for domain generalization.
Moreover, our proposed framework is flexible and can be migrated to various network backbones.
\\
\\
\textbf{Acknowledgments.} We thank anonymous reviewers for the comments and suggestions.
The work described in this paper was supported in parts by the following grants:
Key-Area Research and Development Program of Guangdong Province, China (2020B010165004),
Hong Kong Innovation and Technology Fund (Project No. ITS/426/17FP and ITS/311/18FP),
and National Natural Science Foundation of China with Project No. U1813204.
	
	%
	
	%
	%
	\bibliographystyle{splncs04}
	\bibliography{egbib}
\end{document}